\documentclass[lettersize,journal]{IEEEtran}
\usepackage{amsmath,amsfonts}
\usepackage{algorithmic}
\usepackage{algorithm}
\usepackage{array}
\usepackage[caption=false,font=normalsize,labelfont=sf,textfont=sf]{subfig}
\usepackage{textcomp}
\usepackage{stfloats}
\usepackage{url}
\usepackage{verbatim}
\usepackage{graphicx}
\usepackage{cite}
\usepackage{hyperref}
\usepackage{multirow} 
\begin{document}

\title{A Benchmarking Study of Vision-based Robotic Grasping Algorithms}

\author{Bharath~K~Rameshbabu$^*$,~Sumukh~S~Balakrishna$^*$,~Brian~Flynn$^+$, Vinayak~Kapoor$^*$, ~Adam~Norton$^+$,~Holly~Yanco$^+$,~and~Berk~Calli$^*$
\thanks{This paper was supported in part by the National Science Foundation under awards ECCS-2338703, TI-2229577, TI-2346069, and CNS-1925604}
\thanks{$^*$The authors are with the Robotics Engineering Department, Worcester~Polytechnic~Institute, Worcester, MA 01609, US.}
\thanks{$^+$The authors are with the New England Robotics Validation and Experimentation (NERVE) Center, University of Massachusetts Lowell, Lowell, MA, 01852, US.}
\thanks{E-mail: $\mathtt{\{brameshbabu, ssreenivasarao, vkapoor, bcalli\}@wpi.edu}$\newline$\mathtt{\{Brian\_Flynn, Adam\_Norton\,Holly\_Yanco\}@uml.edu}$}}

\markboth{IEEE Robotics and Automation Magazine}%
{Shell \MakeLowercase{\textit{et al.}}: A Sample Article Using IEEEtran.cls for IEEE Journals}


\maketitle
\begin{abstract}
We present a benchmarking study of vision-based robotic grasping algorithms and provide a comparative analysis. In particular, we compare two machine-learning-based and two analytical algorithms using an existing benchmarking protocol from the literature and determine the algorithms' strengths and weaknesses under different experimental conditions. These conditions include variations in lighting, background textures, cameras with different noise levels, and grippers. We also run analogous experiments in simulations and with real robots and present the discrepancies. Some experiments are also run in two different laboratories using the same protocols to further analyze the repeatability of our results. We believe that this study, comprising 5040 experiments, provides important insights into the role and challenges of systematic experimentation in robotic manipulation and guides the development of new algorithms by considering the factors that could impact the performance. The experiment recordings\footnote{ \href{https://tinyurl.com/5b6sxu22}{https://tinyurl.com/5b6sxu22}} and our benchmarking software\footnote{ \href{https://github.com/vinayakkapoor/vision_based_grasping_benchmarking}{https://github.com/vinayakkapoor/vision\_based\_grasping\_benchmarking}} are publicly available. 
\end{abstract}


\section{Introduction}
\IEEEPARstart{G}{rasping} plays a pivotal role in robotics, enabling machines to interact with and manipulate the objects around them. Such capabilities are indispensable in various applications requiring object pick-and-place, including assisted living and manufacturing robots. The past decade has witnessed a surge in the development of diverse grasping methodologies and gripper designs. Despite these advancements, the lack of standardized benchmarking studies for assessing grasping performance makes it challenging to compare the effectiveness of different grasping systems, evaluate the impact of different parameters or design choices, and systematically test and improve grasp synthesis algorithms. 

In the literature, the lack of utilizing common experimental protocols, experimental setups, and performance metrics pose significant challenges in benchmarking grasping algorithms. Researchers generally apply their own success metrics and evaluation protocols, which makes it difficult to conduct systematic comparisons between methods or to reproduce the results in the research papers. This phenomenon significantly degrades the ``science" aspect of robotics.

At the same time, it is important to recognize the significant and clear challenges to achieving systematic studies in robotic manipulation. There are many factors that impact the performance of a grasping system, which can be difficult to control between different laboratories and studies. The availability (or preference) of hardware, including grippers and cameras, provides variability between experiments. Similarly, utilized objects, uncontrolled lighting conditions and glare, background color, and textures may all influence the experimental results in major ways, and it might be difficult to align the experiments conducted in different research labs on these aspects. 
\newline \indent
Recognizing the benefits of performance benchmarking as well as the abovementioned challenges, this work presents a benchmarking study focusing on vision-based grasp synthesis algorithms that use top-down camera viewpoints. We conducted a comprehensive set of real-world experiments with four different vision-based grasping algorithms based on utilizing a previously published benchmarking protocol \cite{bib3}. We chose the objects from the YCB object set \cite{bib12} so that the experiments are repeatable in other laboratories. In particular, we evaluate two machine-learning-based algorithms \cite{bib7}\cite{bib8} and two analytical algorithms \cite{bib10}\cite{bib11} with distinct approaches in various experimental conditions. These conditions include two different lighting settings, two background textures, two robot manipulators, two grippers, and two cameras. We also run some of our experiments in two different laboratories with the same experimentation protocol and compared the results for a repeatability analysis. While the main focus of this paper is real-world experiments, we also ran several analogous simulation tests to analyze the discrepancies between the real-world settings and simulations. We provide the code for applying the benchmarking protocol \cite{bib3} to streamline future benchmarking efforts. Additionally, we provide tools for both data collection and visualization of the results. 
\newline \indent
It is important to note that none of these algorithms was developed in the authors' laboratories. An extensive effort is spent to utilize the algorithms in their full capacity and allow for a fair analysis. Our goal was to achieve an apples-to-apples comparison as much as possible and discuss the results along with the challenges of running a comparative experimental study in robotic manipulation. We believe that the analysis helps identify the failure modes of the examined grasping algorithms and lays the groundwork for the developers to apply the same experimental protocol and directly compare their new algorithms with the analyzed ones without the laborious work of repeating the experiments of the prior work.

\section{Related Work}
In the robotic grasping literature, performance analysis is generally conducted either by using datasets or by conducting experiments via custom protocols of individual laboratories. While there are valuable datasets that can be used for designing and testing grasp synthesis algorithms such as the Jacquard \cite{bib1} and the Cornell \cite{bib2}, their performance evaluation metric is primarily based on the `intersection over union’ criterion (IoU), which does not sufficiently reflect the success of the algorithms in real world applications \cite{bib8}\cite{bib23}. GraspNet-1Billion \cite{bib14}, on the other hand, is a large-scale benchmark for general object grasping that provides over one billion 3D grasp data, utilizing a force closure-based evaluation metric. Jacquard \cite{bib1} provides a web interface for researchers to evaluate the performance of grasps in a simulated environment. Similarly, DEX-NET \cite{bib15} provides a dataset of synthetic objects and grasps based and metrics based on GWS (Grasp Wrench Space). VisGrab \cite{bib5} provides tooling to evaluate algorithms only in a simulated robot setup. However, despite all the progress, the discrepancies between the datasets, simulated experiments, and the real world experiments prevent an accurate analysis \cite{bib3}\cite{bib16}\cite{bib5}\cite{bib6}. Therefore, the robotics community generally expects grasping algorithms to be assessed through real world experiments (e.g., when they are presented in research papers).
\newline \indent
Some notable contributions have been made in standardizing benchmarking experiments for robotic grasping, such as the protocols proposed by Bekiroglu et al. \cite{bib3} and GRASPA 1.0 \cite{bib16}. VisGrab \cite{bib5} and the work by Vuuren et al. \cite{bib6} propose evaluation metrics and benchmarking procedures that focus on vision-based grasping algorithms. Vuuren et al. \cite{bib6} focus on a specific application of evaluating algorithms for objects on a conveyor belt setup. Bottarel et al. \cite{bib16} conducted a comparative analysis of three data-driven algorithms under a single experimental setup using the GRASPA 1.0 benchmarking protocol, aligning closely with the focus of our present work in this paper. In an effort to span other prominent approaches in the grasping literature that utilize a top-down camera viewpoint and assess the conditions that impact the algotrithm performance, we conduct a benchmarking study that includes a detailed analysis of the performance and failure modes of four different algorithms (two data-driven and two analytical) across 10 different objects in diverse experimental conditions and robot configurations. In this work, we utilized the benchmarking protocol in \cite{bib3}, which is the most suitable one for our case study, i.e., single object grasping with top-down camera viewpoint; other protocols either focus on 6D grasping with different viewpoints or primarily on multi-object settings. In selecting the algorithms for our study, we drew insights from three pivotal survey papers \cite{bib17}\cite{bib19}\cite{bib25} in the field of robotic grasping. We prioritized data-driven algorithms with easy-to-adopt open source implementations, while we modernized analytical algorithms to demonstrate a baseline comparison.

\section{The Algorithms}
\begin{figure}[t]
  \begin{center}
  \includegraphics[width=0.48\textwidth]{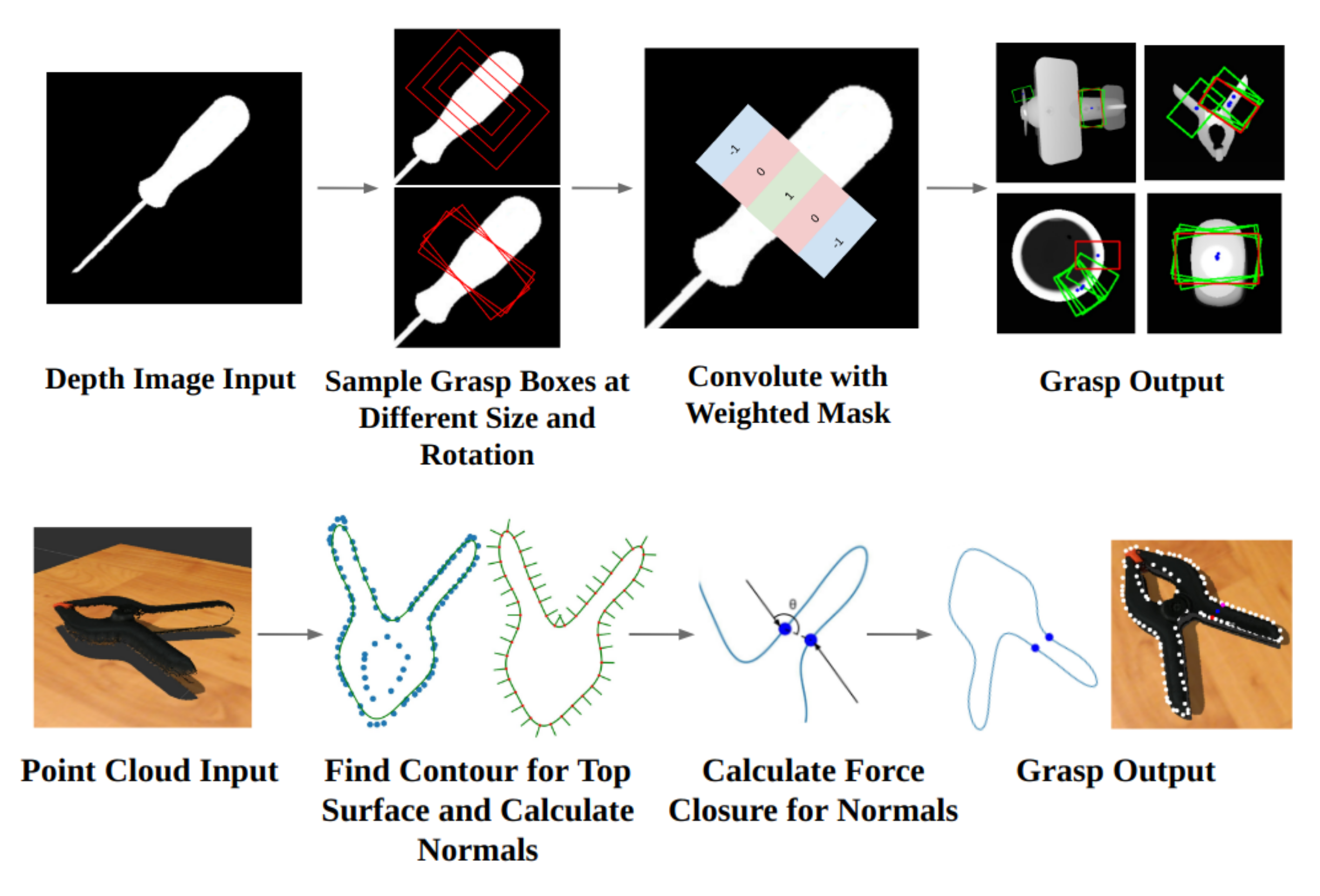}
  \end{center}
  \captionsetup{font=footnotesize}
  \vspace{-10pt}
  \caption{Illustrations of the flow of analytical algorithms. The image on top represents the mask-based algorithm, and the image on the bottom represents the top surface algorithm.}
  \label{fig:top}
\end{figure}

\begin{figure*}[t]
  \begin{center}
  \includegraphics[width=0.96\textwidth]{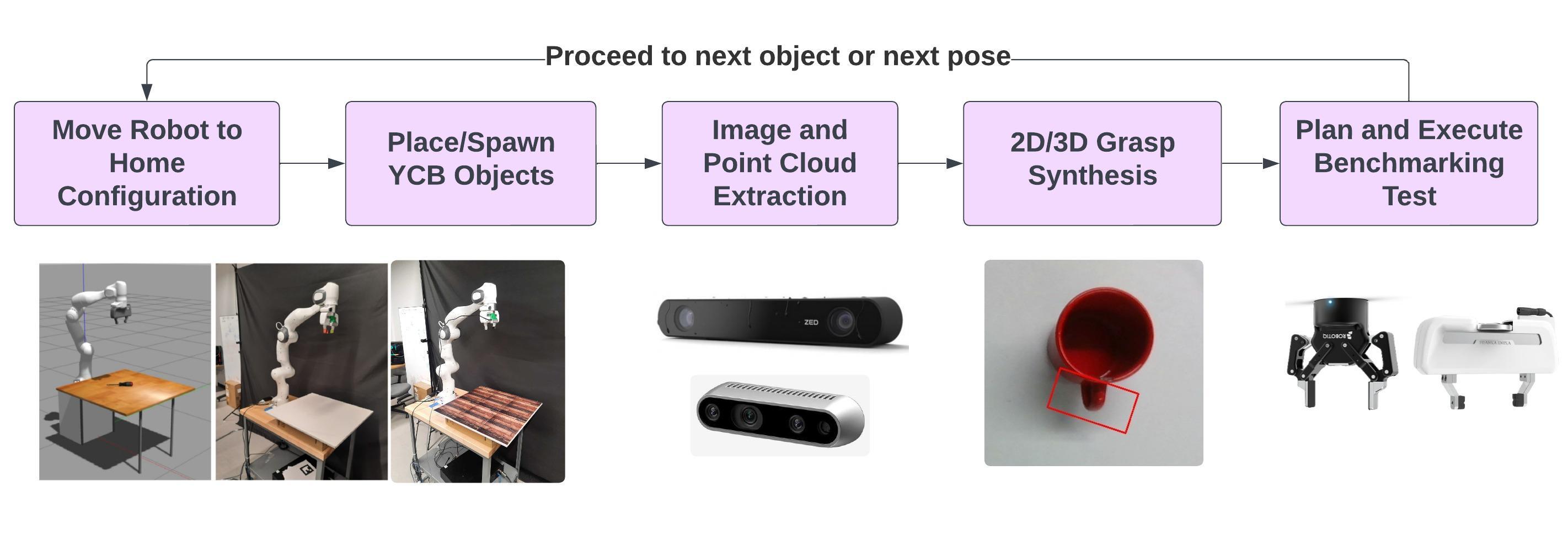}
  \end{center}
  \captionsetup{font=footnotesize}
  \vspace{-10pt}
  \caption{The diagram describes the workflow of the benchmarking pipeline. The images at the bottom show different configurations used for the experiment. The left image shows Franka Panda in the simulation and real world environment with textured and non-textured backgrounds. The experiments utilized Intel RealSense and ZED 2i cameras for perception and Franka and Robotiq grippers for grasping. }
  \label{fig:benchmarking_workflow}
\end{figure*}

We selected four grasping algorithms that utilize a top-down image of the scene. Two of these algorithms are data-driven, while the other two are based on classical approaches. We perform a set of preprocessing steps to enhance the input data. For the depth image of the RealSense camera, we apply KDTree-based depth completion and fill the holes with the nearest available depth. For the analytic algorithms that utilize point clouds, we filtered out the major plane. To all the algorithms, we applied ROI filtering to remove the unwanted information outside our region of interest.

\subsection{Data-driven Algorithms}
We evaluated two learning-based algorithms, i.e., GG-CNN \cite{bib7} and ResNet-based \cite{bib8}. The implementation of the learning-based algorithms was directly adopted from the public repositories of the papers. GG-CNN utilizes an FCN-based model to generate dense maps for grasp quality, angle, and grasp width using a depth image as input. On the other hand, the ResNet-based model integrates components from Mask R-CNN with ResNet to generate grasp rectangles using RGD input, where the green channel is replaced with a depth image.

\subsection{Analytical Algorithms}
The analytical algorithms are based on \cite{bib10}, which segments and utilizes the top surfaces of the objects, and \cite{bib11}, which adopts a
mask-based approach to determine the grasp poses. The analytical approaches were developed in 2008 and 2011, when computational power was comparatively limited. Therefore, these methods choose to use some heuristics while searching for grasp candidates. In our work, we took advantage of today’s improved computational resources and enhanced the implementation of these algorithms (e.g., allowing the algorithms to exhaustively search grasps instead of using heuristics) to have a fair comparison with the performance of the data-driven algorithms.
The overview of the top surface algorithm is depicted in Figure \ref{fig:top}. The top surface-based algorithm \cite{bib10}\cite{bib9} segments the point cloud of the object and distance-filters the top layer (e.g., 2 cm-thick region) of the resulting point cloud \cite{bib10}. This portion is projected to a 2D plane along the table normal, and a 2D force closure grasp is searched along its boundaries. Here, instead of using the original paper’s heuristic-based grasp search approach (which was designed to reduce the computational complexity), we apply a brute-force grasp search on the boundary data: First, we determine the concave hull of the top surface. We compute and fit an Elliptical Fourier Descriptor to the concave hull \cite{bib9}. Then, we sample points evenly along the fitted curve and calculate normals along the sampled points by double differentiation. Finally, we calculate force/moment balance \cite{bib24} across densely sampled normals to identify the most stable pair of normals as the candidate grasp. 
The overview of the mask-based algorithm is depicted in Fig. \ref{fig:top}. The mask-based algorithm \cite{bib11} relies on a scoring function that calculates spatial histograms on the sampled convolution masks. Similarly, we designed a weighted mask to search for geometric features on the object that closely resemble the gripper’s structure. The weights in the mask are distributed such that object segments that are located at the center of the grasp are rewarded, while those on the edges are penalized, as these positions correspond to the gripper finger locations. Multiple masks of different sizes and discretized rotations are sampled and convoluted across the entire depth image. Each mask in the convoluted image represents a similarity score, and the mask with the best similarity score is chosen as the candidate grasp. While the original algorithm does relatively sparse sampling and requires a grasp pose refinement step, we skip this step by applying dense sampling.

\section{The Benchmarking Pipeline}
We developed a benchmarking software designed to evaluate the performance of grasping algorithms, both in simulation
and real-world systems, based on the benchmarking protocol from \cite{bib3} (Fig. \ref{fig:benchmarking_workflow}). The software was built using ROS as its backend. We make use of Gazebo for simulation and RViz for grasp visualization. The high-level flow of the benchmarking pipeline is as follows. On launching the software, the robot moves to a predefined home configuration where it takes the top-down images of the scene. As outlined in \cite{bib3}, the benchmarking protocol specifies placing the object at 6 different positions and orientations based on parameters \textit{r} representing the radius of the circle and $\theta$ representing the orientation of the object (r = 25cm and $\theta$ = 90deg for our experiments). The software prompts the user to place the objects to one of these predefined poses. In the case of simulation, the object is automatically spawned at the determined pose. 

Following that, the camera captures RGB and depth information of the scene, which is then subjected to preprocessing. The preprocessing steps include depth completion, point cloud downsampling, and ROI filtering. The grasp synthesis algorithms take in the preprocessed input and produce a candidate grasp. The synthesized grasp is converted into gripper poses by applying rigid body transformations based on the camera and robot parameters (intrinsics and extrinsics). 

The motion planning module performs a linear trajectory to the desired gripper pose and executes the detected grasp. Once grasped, the robot tests the grasp stability (as explained in \cite{bib3} and in the next section) by shaking it and rotating the object to various orientations. Following the benchmarking approach in \cite{bib3}, the target circle's position and orientation ($\theta$) are parameterized in the world frame at the robot's base. The grasp score is assigned based on whether the object remains grasped by the gripper at each stage. The grasp score is automatically calculated in the simulation, while for the real-world experiment, it has to be manually noted and calculated. The robot then returns to its home position, and the process continues with the next object or the next object pose. By allowing adjustments to the target's pose and the robot's initial view, our framework can systematically evaluate grasping algorithms in various workspace configurations.

The software is designed with a high degree of modularity, with each step functioning as a separate ROS node. The configuration file allows for easy parameter adjustments, enabling relatively straightforward customization and accommodating different robots, grippers, cameras, and preprocessing steps with reasonable ease. This flexibility makes it highly user-friendly, facilitating quick adjustments on the go. The simulation is readily accessible, featuring Intel RealSense, Franka Panda robot, a Panda gripper, and a base configuration for convenient testing of grasping algorithms. The algorithms can be easily integrated into the pipeline as a ROS node, and to aid in this process, a template is included in the code base.

\section{Experiments}
\begin{figure*}[htbp]
  \begin{center}
  \includegraphics[width=1\textwidth]{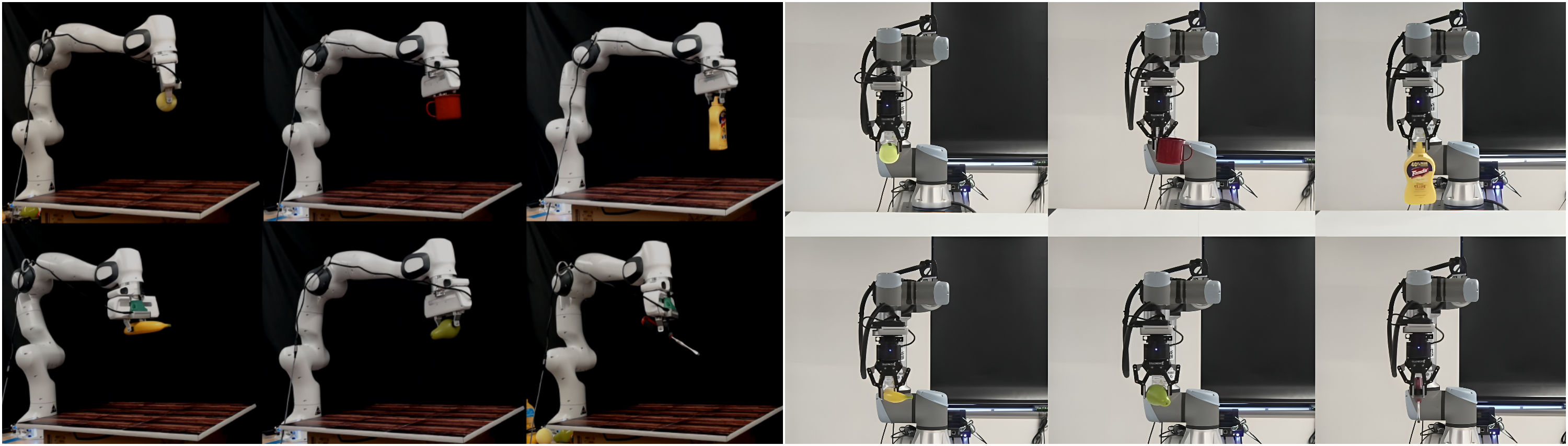}
  \end{center}
 \captionsetup{font=footnotesize}
  \caption{The images on the left show the Franka Panda robot equipped with a Franka Gripper performing object-picking tasks in our lab at Worcester Polytechnic Institute (WPI), operating on a textured grasping surface under ambient lighting of 60 lux. The images on the right show a Universal Robot with a Robotiq Gripper picking objects in our lab at University of Massachusetts Lowell, operating on a non-textured grasping surface under ambient lighting of 340 lux.}
  \label{fig:grasps_panda}
\end{figure*}

\subsection{Experiment Setup}
We conducted experiments with various different hardware. As for robots, we utilized Franka Emika Panda and Universal Robot 5e with Franka Gripper and Robotiq 2F-85 respectively (Fig.~\ref{fig:grasps_panda}). We conducted experiments with two different cameras in an eye-in-hand configuration: Intel RealSense D435 and ZED 2i. For both cameras, we set the exposure to 8500ms, gain to 16db and laser power to 150 mW. The cameras view the grasping scene directly from the top, from a distance of 0.8 meters from the table plane. All robot movements were executed at a joint velocity of 1.0175 rad/s. The lights in the room were positioned on both sides of the setup to minimize the shadows cast by the robot on or near the placed object.

\begin{figure}[htbp]
  \centering
  \hspace*{0.0cm}
  \begin{minipage}[b]{0.48\textwidth}
    \centering
    \includegraphics[width=\textwidth]{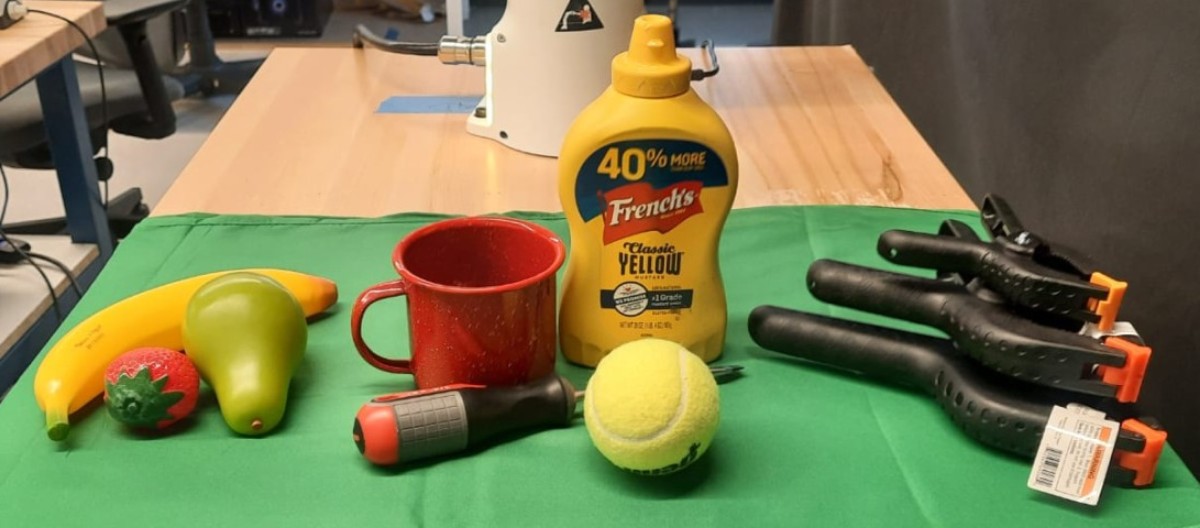}
    \vfill
  \end{minipage}
  \captionsetup{font=footnotesize}
 \caption{The YCB objects utilized for the benchmarking experiments.}
 \label{fig:obj}
\end{figure}

To conduct our experiments, we selected 10 objects from the YCB object dataset \cite{bib12} depicted in Fig. \ref{fig:obj} and Table \ref{tab:baseline-additional-comparison}. The selection of objects was curated to encompass diverse shapes, weights, colors, and frictional properties. Among the chosen items, the banana, pear, and strawberry were selected for their predominantly curvy surfaces and distinctive colors. The mug, mustard bottle, screwdriver, and tennis ball were chosen for their varied shapes, while considering their common usage in manipulation applications. The inclusion of three clamp objects, differing only in size, enabled an evaluation of the algorithms' size variance.

\subsection{Benchmarking Protocol and Scoring}
\begin{algorithm}
\footnotesize
\caption{Scoring For a Single Algorithm}
\label{alg:alg1}
\begin{algorithmic}[1]
    \STATE Initialize Points Scored = 0
    \FOR {each object}
        \FOR {each pose}
            \STATE Synthesize grasp for the object at the given pose
            \STATE Pick up the object and stop at 40 cm vertically.
            \IF {Object is still grasped}
                \STATE Points Scored += 1
            \ENDIF
            \STATE Yaw the object to +/- 45 degrees
            \IF {Object is still grasped}
                \STATE Points Scored += 1
            \ENDIF
            \STATE Shake the object by pitching +/- 45 degrees
            \IF {Object is still grasped}
                \STATE Points Scored += 1
            \ENDIF
        \ENDFOR
    \ENDFOR
    \STATE Grasp Score = Points Scored / Total Possible Points
\end{algorithmic}
\end{algorithm}

To evaluate and compare different grasping algorithms, we adopted the benchmarking procedure described in \cite{bib3}. We conducted the experiments in six different (but consistent) locations for each object. In each location, we attempted a pick for the object based on the output of the grasping algorithm. If the pick is successful, we perform yaw movements of +/- 45 degrees and shaking motions with an angle of +/- 45 degrees to test grasp stability. Our scoring methodology comprises a grasp success score and a grasp score assigned to each object. The grasp success score denotes the success or failure of the object's grasp, while the grasp score is a cumulative measure of points after each stage of the benchmarking protocol, reflecting the stability of the grasp and the impact of object properties on the grasping performance. The scoring strategy for a single algorithm is described in Algorithm 1.

\subsection{Experiment Configurations}
\begin{table}[t]
  \caption{Different experiment configurations that are considered for this benchmarking study.}
  \label{tab:baseline-additional-comparison}
  \centering
  \begin{tabular}{ll}
    \hline
    \\[-1.8ex]
    \textbf{Category} & \textbf{Experiment Configurations} \\
    \\[-1.8ex]
    \hline
    \\[-1.0ex]
    Algorithms & \begin{tabular}[t]{@{}p{0.6\linewidth}@{}}GG-CNN, ResNet, Top Surface, Mask-based \end{tabular} \\
    \\[-1.8ex]
    Cameras & \begin{tabular}[t]{@{}p{0.6\linewidth}@{}}Intel Realsense D435, ZED 2i \end{tabular} \\
    \\[-1.8ex]
    Background & \begin{tabular}[t]{@{}p{0.6\linewidth}@{}}Textured, Non Textured \end{tabular} \\
    \\[-1.8ex] 
    Lighting Condition & \begin{tabular}[t]{@{}p{0.6\linewidth}@{}}60 Lux, 340 Lux \end{tabular} \\ 
    \\[-1.8ex]
    Manipulator & \begin{tabular}[t]{@{}p{0.6\linewidth}@{}}Franka Emika Panda, Universal Robot 5e \end{tabular} \\
    \\[-1.8ex]
    Grippers & \begin{tabular}[t]{@{}p{0.6\linewidth}@{}}Franka Emika Gripper, Robotiq 2F-85 \end{tabular} \\
    \\[-1.8ex]
    Environment & \begin{tabular}[t]{@{}p{0.6\linewidth}@{}}Real World Setup, Gazebo Simulation \end{tabular} \\
    \\[-1.8ex]
    YCB Objects & \begin{tabular}[t]{@{}p{0.6\linewidth}@{}}Tennis Ball, Screw Driver, Medium Clamp, Large Clamp, Extra Large Clamp, Banana, Strawberry, Pear, Mug, Mustard \end{tabular} \\
    \\[-1.0ex]
    \hline
  \end{tabular}
\end{table}

\begin{table*}[htbp]
\centering
\caption{Grasping Performance}
\label{tab:results}
\renewcommand{\arraystretch}{1.5}
\begin{tabular}{p{2.7cm}cccccccccc}
\hline  
\textbf{Configuration} & \textbf{Background} & \textbf{Lighting} & \multicolumn{2}{c}{\textbf{GG-CNN}} & \multicolumn{2}{c}{\textbf{ResNet}} & \multicolumn{2}{c}{\textbf{Top Surface}} & \multicolumn{2}{c}{\textbf{Mask Based}} \\ 
\cline{4-11}   &  &  & \textbf{GSS} & \textbf{GS} & \textbf{GSS} & \textbf{GS} & \textbf{GSS} & \textbf{GS} & \textbf{GSS} & \textbf{GS} \\     \hline   
Simulation & Non Textured & 340 lux & 0.567 & 0.567 & 0.750 & 0.750 & \textbf{0.883} & \textbf{0.883} & 0.850 & 0.850 \\ \hline  
\multirow{4}{=}{Franka Panda, Franka Gripper, Intel Realsense (WPI)} & Textured & 340 lux & 0.250 & 0.255 & 0.250 & 0.255 & 0.816 & 0.816 & \textbf{0.883} & \textbf{0.883} \\ \cline{3-11}  
 &  & 60 lux & 0.350 & 0.350 & 0.250 & 0.255 & 0.916 & 0.922 & \textbf{0.950} & \textbf{0.961} \\ \cline{2-11}  
 & Non Textured & 340 lux & 0.266 & 0.277 & 0.566 & 0.572 & 0.766 & 0.777 & \textbf{0.850} & \textbf{0.861} \\ \cline{3-11}  
 &  & 60 lux & 0.316 & 0.327 & 0.716 & 0.716 & 0.900 & 0.922 & \textbf{0.916} & \textbf{0.927} \\  \hline  
\multirow{4}{=}{UR5, Robotiq Gripper, Intel Realsense (UML)} & Textured & 340 lux & 0.450 & 0.450 & 0.267 & 0.278 & \textbf{0.966} & \textbf{0.966} & 0.883 & 0.883 \\ \cline{3-11}  
 &  & 60 lux & 0.400 & 0.400 & 0.333 & 0.333 & \textbf{0.966} & \textbf{0.966} & 0.883 & 0.888 \\ \cline{2-11}  
 & Non Textured & 340 lux & 0.433 & 0.433 & 0.700 & 0.700 & \textbf{0.966} & \textbf{0.966} & 0.950 & 0.950 \\ \cline{3-11}  
 &  & 60 lux & 0.350 & 0.355 & 0.750 & 0.750 & \textbf{0.966} & \textbf{0.966} & 0.900 & 0.900 \\ \hline  
\raggedright{Franka Panda, Robotiq Gripper, Intel Realsense (WPI)} & Non Textured & 340 lux & 0.355 & 0.355 & 0.572 & 0.572 & 0.800 & 0.800 & \textbf{0.855} & \textbf{0.855} \\ \hline  
\raggedright{Franka Panda, Franka Gripper, ZED 2i (WPI)} & Non Textured & 340 lux & 0.410 & 0.410 & 0.700 & 0.700 & 0.816 & 0.816 & \textbf{0.850} & \textbf{0.850} \\ \hline 
\end{tabular}

\vspace{0.2cm}
\raggedright
The table presents the grasping Performance for 10 YCB objects from Figure \ref{fig:obj}.  Grasp Score (GS) is scored based on Algorithm \ref{alg:alg1}, which takes into account partial scoring. Grasp Success Score (GSS), on the other hand, quantifies binary success scores from each experiment. The maximum possible score for each table cell is 1. (UML) indicates the corresponding experiments are conducted at the University of Massachusetts Lowell.
\end{table*}

To thoroughly assess the performance of the grasping algorithms, we conducted evaluations and comparisons in diverse experimental conditions as summarized in Table \ref{tab:baseline-additional-comparison}. The algorithms were tested in both simulated and real-world environments to evaluate
their effectiveness in different contexts. We systematically varied various experiment parameters to assess and compare the performance of the algorithms in different conditions as follows: We tested the algorithms in the two
different background textures as shown in Fig.~\ref{fig:benchmarking_workflow} under varying lighting conditions (60 lux and 340 lux) to determine
their robustness in real-world scenarios. We conducted experiments with two parallel jaw grippers: the Franka gripper and the Robotiq gripper.  Furthermore, we utilized two different cameras: Realsense D435 and ZED 2i. The depth from the RealSense camera was susceptible to noise and influenced by changes in lighting, while the ZED 2i camera incorporated an internal ML algorithm for precise depth prediction. In addition, we conducted some of the experiments in two different laboratories under the supervision of different researchers to assess the reproducibility of our results.
In total, we conducted 5040 distinct experiments, considering each instance of testing a single object grasp as an individual experiment. This encompasses 3 trials, 10 objects, 6 poses, 2 background textures, 2 lighting conditions, 2 cameras, 2 grippers, 2 manipulators, and 4 algorithms.

\section{Results}
\begin{figure}[t]
  \begin{center}
  \includegraphics[width=0.45\textwidth]{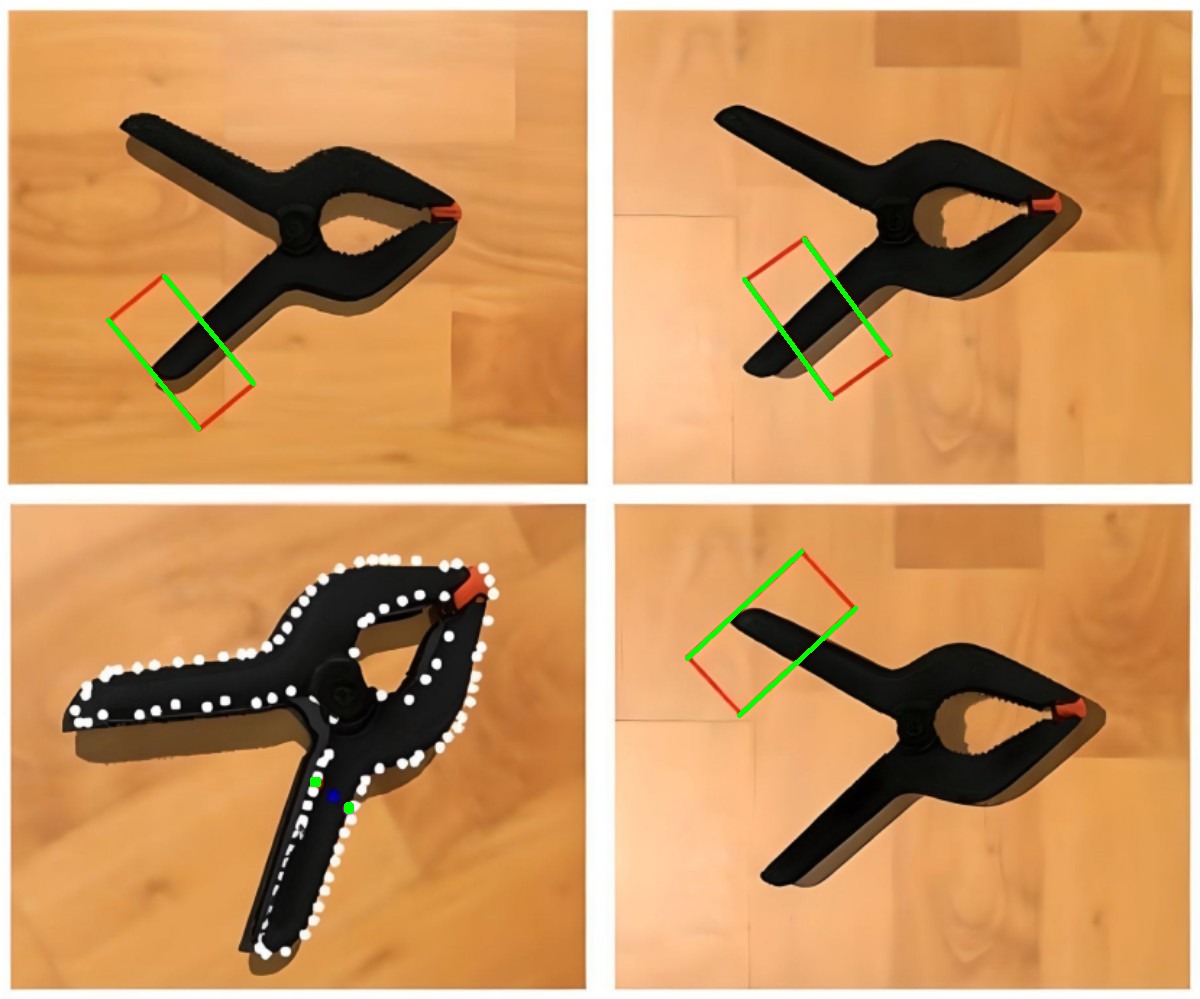}
  \end{center}
 \captionsetup{font=footnotesize}
  \caption{Grasps synthesized  by different grasping algorithms for Medium Clamp in the simulation setup (Mask-based, ResNet, Top Surface, GG-CNN in order from top-left to bottom-right.)}
  \label{fig:grasps}
\end{figure}
Table \ref{tab:results} shows the grasping performance of 10 YCB objects (Fig. \ref{fig:obj}). The table includes both Grasp Score (GS) and Grasp Success Score (GSS) for all four algorithms in different experiment configurations (Table \ref{tab:baseline-additional-comparison}). Each entry of the table represents 60 experiments (10 objects in 6 different locations). We made video recordings of the experiments publicly available to provide deeper insights into the performance of the algorithms in the provided link\footnote{ \href{https://tinyurl.com/5b6sxu22}{https://tinyurl.com/5b6sxu22}}. The table also indicates the locations of each experiment set i.e., whether the results are collected at MER Lab (Worcester Polytechnic Institute; WPI) or NERVE Lab (University of Massachusetts Lowell; UML).

In addition, we provide some selective results that offer interesting insights into how different configurations impact the performance of the algorithms. 
\begin{itemize}
    \item The top-left plot in Fig. \ref{fig:ggcnn-sim-real} shows a comparison of the grasping performance of GG-CNN in the simulation versus the real-world setting. 
    \item The top-right plot in Fig. \ref{fig:ggcnn-sim-real} compares the performance of ResNet in textured and non-textured backgrounds. 
    \item The bottom plots in Fig. \ref{fig:ggcnn-sim-real} illustrate the impact of varying lighting conditions on the Top Surface and the ResNet algorithm. 
    \item Fig. \ref{fig:camera-comp} presents the grasping performance of GG-CNN and ResNet with Intel RealSense and ZED 2i cameras. 
    \item The top plot in Fig. \ref{fig:resnet-pose} illustrates the grasping performance of ResNet at each position from the benchmarking protocol \cite{bib3}, highlighting the effects of lighting conditions.
    \item The middle plot in Fig. \ref{fig:resnet-pose} illustrates the performance of the algorithms at different laboratories.
    \item The bottom plot in Fig. \ref{fig:resnet-pose} shows the performance of GG-CCN with different grippers (Robotiq 2F-85 and Franka Gripper). 

\end{itemize}
These results are discussed below.

\subsection{Analytical Algorithms vs. Data-driven Algorithms}
In Table II, we present the grasping performance of all four algorithms in various experiment configurations. The results indicate that the analytical algorithms, specifically the adapted versions of the Mask-based and the Top Surface algorithm, exhibit better performance compared to the learning algorithms for the majority of the objects, with scores consistently above 0.8 (out of a maximum score of 1) in all the environment configurations. 
\begin{figure*}[t]
  \begin{center}
  \includegraphics[width=1\textwidth]{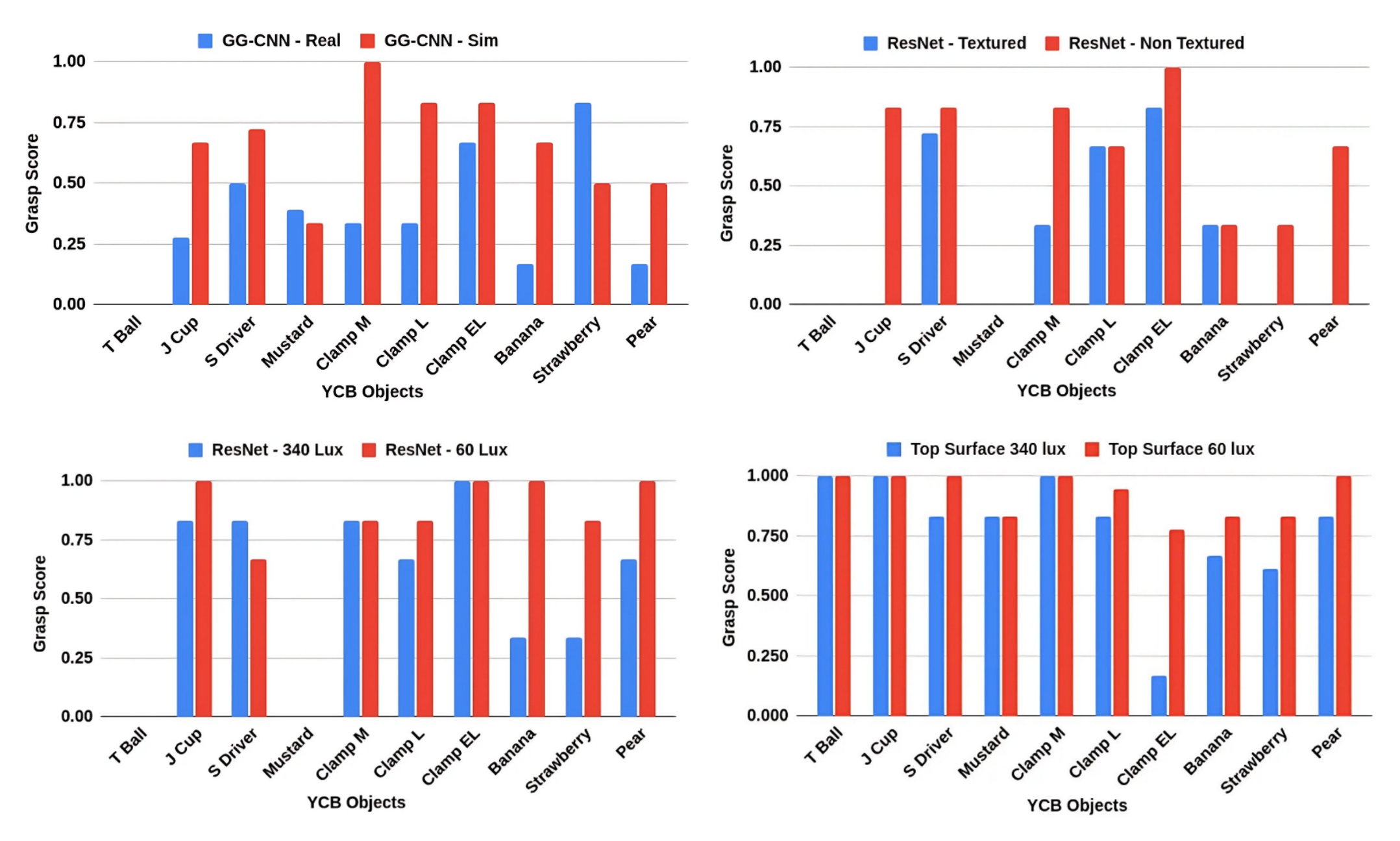}
  \end{center}
  \captionsetup{font=footnotesize}
  \vspace{-10pt}
  \caption{The top-left plot shows the grasping performance of GG-CNN in simulation and real-world setup. The top-right plot shows the grasping performance of ResNet in Textured and Non-Textured background (Franka Panda, Franka Gripper, and Intel RealSense configuration). The bottom-left plot shows the grasping performance of ResNet in 340 lux and 60 lux brightness (Franka Panda, Franka Gripper, and Intel RealSense configuration). The bottom-right plot shows the grasping performance of Top Surface in 340 lux and 60 lux brightness (Franka Panda, Franka Gripper, and Intel RealSense configuration).}
  \label{fig:ggcnn-sim-real}
\end{figure*}

On the other hand, the learning algorithms, GG-CNN and ResNet, exhibited inconsistent performance across different experiment configurations. They perform exceptionally well for a specific set of
objects, but demonstrate poor performance for some other objects. This discrepancy can primarily be attributed to the specific objects the models were trained on, as we utilized pretrained models for our evaluation. Similarly, these algorithms are observed to be vulnerable to changes in lighting conditions and background textures, likely due to the limited variations of these factors in the specific datasets on which they were trained. In the simulation, the conditions are noise-free, allowing the learning-based algorithms to perform well, leading to relatively higher scores. This can also be clearly observed from the top-left plot in Fig. \ref{fig:ggcnn-sim-real}, which illustrates that GG-CNN performs significantly better for most objects in simulation, indicating that the algorithm is sensitive to real-world environmental factors.

\subsection{Impact of background texture}
Another notable finding is that ResNet excels in performance when the objects are placed against a non-textured background. This is due to ResNet being an RGD (depth replaces blue channel) based algorithm, where the presence of color and edge features in the textured background image significantly impacts its performance. The top-right plot in Fig. \ref{fig:ggcnn-sim-real} shows how ResNet performs better for all the objects in a non-textured background with relatively higher grasp scores compared to the textured case. Other algorithms performed more or less the same on different background textures since they only use depth information as input.

\begin{figure}[t]
  \begin{center}
  \includegraphics[width=0.44\textwidth]{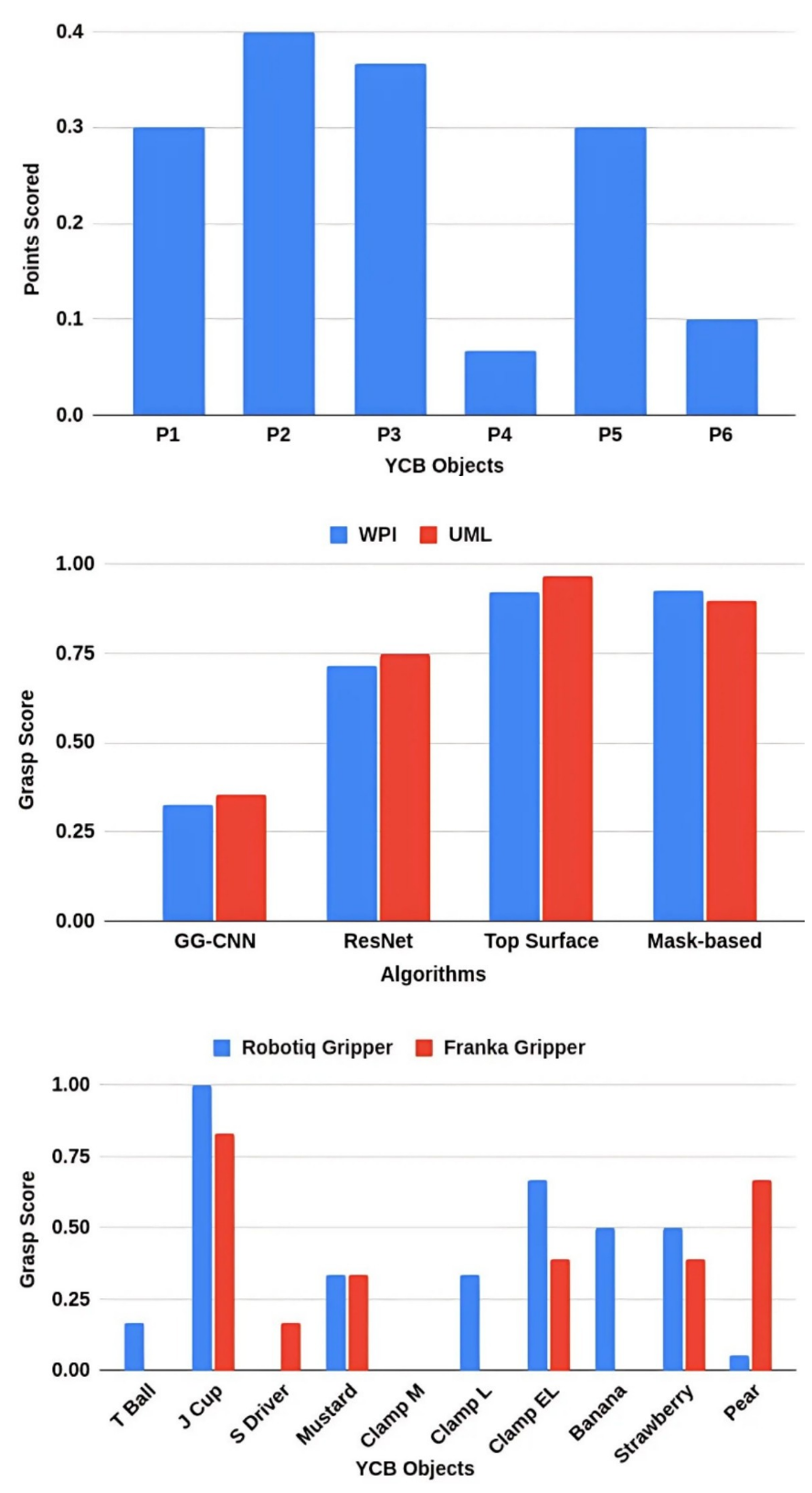}
  \end{center}
  \captionsetup{font=footnotesize}
  \vspace{-10pt}
  \caption{The top plot shows the grasping performance of ResNet at 340 lux across 6 positions/orientations from the benchmarking protocol \cite{bib3} (Franka Panda, Franka Gripper, and Intel Realsense configuration). The middle plot compares the grasping performance of all the algorithms at two different labs (Worcester Polytechnic Institute (WPI) and University of Massachusetts Lowell (UML)). Bottom plot compares the grasping performance of GG-CNN with Franka Gripper vs Robotiq gripper (Franka Panda and Intel RealSense).}
  \label{fig:resnet-pose}
\end{figure}

\subsection{Impact of lighting conditions}
ResNet is also significantly affected by glare from surrounding lighting. The top plot in Fig. \ref{fig:resnet-pose} displays the performance of ResNet at 340 lux for the 10 objects in the set at different poses as per the benchmarking protocol from \cite{bib3}. Grasp scores at P4 and P6 (identical positions with different orientations, refer \cite{bib3} for object pose definitions) exhibited poorer performance compared to other positions. The performance decrease is due to glare on that particular position, leading to lower-quality RGB data. Given that ResNet utilizes both RGB and depth data to detect grasps, this degradation in image quality adversely impacted its performance. This performance decrease is not observed for GG-CNN, since it only utilizes depth images.

We also noted that the lighting conditions affected the performance when RealSense is used. The bottom plots in Fig. \ref{fig:ggcnn-sim-real} illustrate the performance of ResNet and Top Surface for the 10 object set at 60 lux and 340 lux lighting conditions. The lower lighting conditions resulted in better performance by the algorithms. This is due to the noisy depth data from Intel RealSense at 340 lux. Even with the integration of a depth completion step in the benchmarking pipeline, the resulting depth quality remained subpar due to the noisy estimation in the depth image. As observed in the bottom plots of Fig. \ref{fig:ggcnn-sim-real}, better point clouds at 60 lux lighting resulted in better performance by the Top Surface algorithm, whereas for ResNet, the absence of glare along with better depth data improved results at 60 lux lighting conditions.      

\begin{figure}[t]
  \begin{center}
  \includegraphics[width=0.5\textwidth]{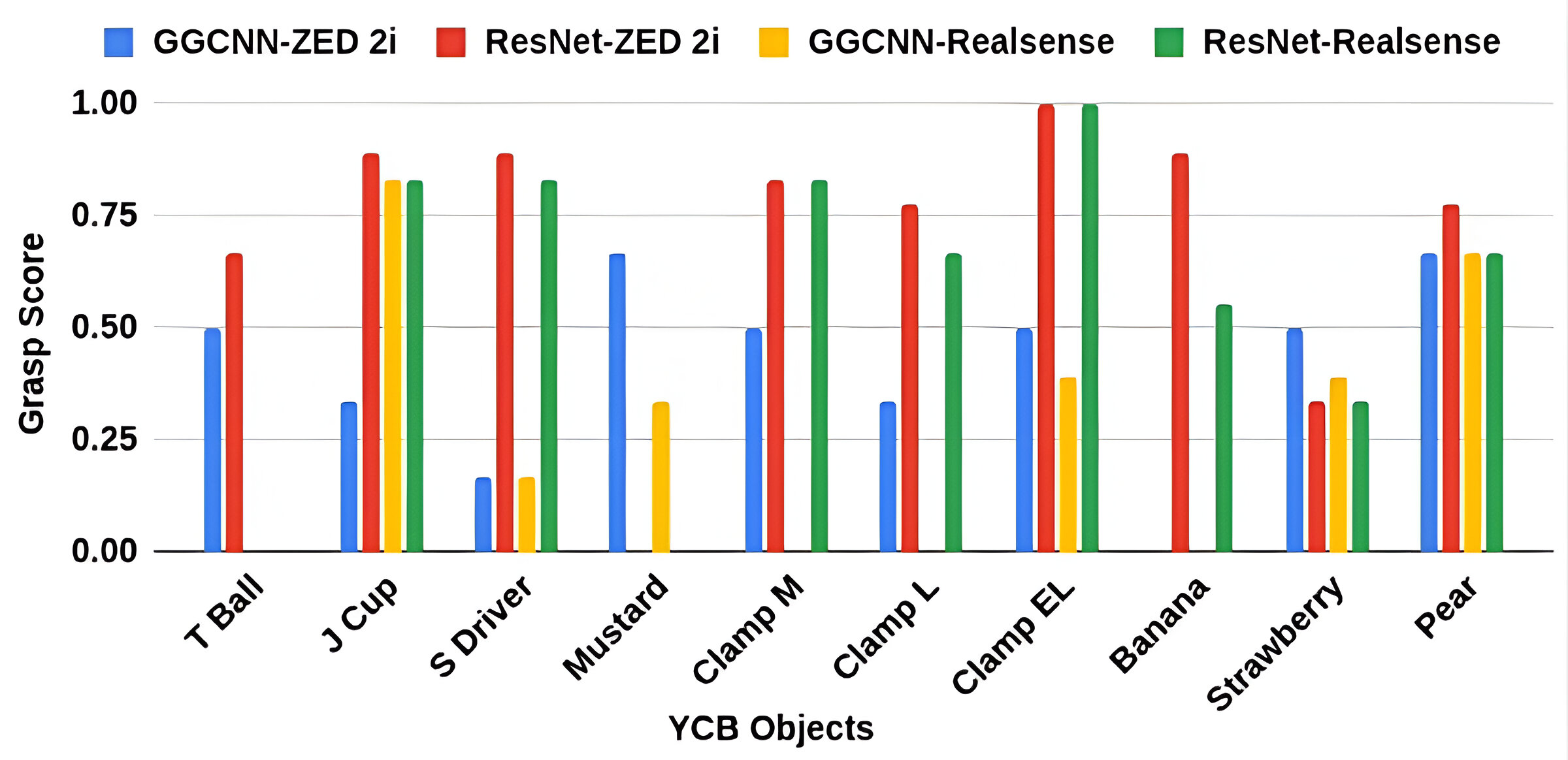}
  \end{center}
  \captionsetup{font=footnotesize}
  \caption{Grasping performance of GG-CNN and ResNet for 10 objects from Figure \ref{fig:obj}, with ZED 2i and Intel Realsense cameras (Franka Panda, Franka Gripper configuration).}
  \label{fig:camera-comp}
\end{figure}

\subsection{Impact of camera selection}

Fig. \ref{fig:camera-comp} shows the improved performance of GG-CNN and ResNet when using the ZED 2i RGBD camera for most of the objects, except a couple of outliers. ZED 2i uses an internal ML algorithm to predict high-quality depth, robust to environmental factors. It is evident that the ZED 2i generally yielded almost equal or higher grasp scores in all cases. In the rare cases of worse performance with ZED 2i, it is possibly due to depth artifacts generated by the internal ML algorithms that don't accurately represent the object. The use of KDTree-based depth completion only helped Intel RealSense experiments slightly as ZED 2i's internal ML algorithm worked significantly better.

\subsection{Identical experiments run in two different labs}
We run identical experiments in two different laboratories and research scientists using the same benchmarking protocols. We detected small disparities in the algorithm performances, and the general score trends were similar as observed in the middle plot of Fig.~\ref{fig:resnet-pose}. The differences can be attributed to the absence of ambient light and the presence of shiny spots in the MER Lab setup, resulting in lower quality depth perception by Intel RealSense, thereby causing a slightly inferior performance of the top surface algorithm. These results demonstrate the consistency of the benchmarking study and the comparative results while also showing the challenges of replicating the exact same results in different locations.

\subsection{Effects of different grippers}
Regarding the impact of gripper hardware, the bottom plot in Fig. \ref{fig:resnet-pose} depicts the comparison of GG-CNN performance utilizing the Franka Gripper and the Robotiq Gripper. The Robotiq Gripper, characterized by its larger finger contact area, exhibited superior performance in contrast to the Franka Gripper, which possesses a relatively smaller contact area. This trend was evident across all algorithms, not solely limited to GG-CNN. The Franka Gripper was less likely to pick up objects for slightly sub-optimal grasp detections compared to the Robotiq Gripper, which showed more tolerance towards such detections.

\subsection{Effects of different object properties}
We observed that the mask-based algorithm has consistent performance across all objects, but mustard due to its difficult-to-grasp top surface. The Top Surface algorithm encountered challenges with smaller objects like the medium clamp, as well as with J Cup, due to inaccuracies in its concave hull calculation for circular hollow objects, which occurred sporadically. ResNet and GG-CNN consistently exhibited lower performance when handling objects with curved surfaces, such as fruits, or objects with challenging top surfaces. The tennis ball also poses a challenge, since successfully grasping it requires a very accurate gripper positioning that is aligned with the ball's diameter; even small inaccuracies may either cause the fingers to push the ball away, or one of the fingers may touch the ball earlier than the other, making it move away from the grasp. We observed that the data-driven strategies struggle to deliver that accuracy, while the analytical approaches provide high performance. GG-CNN exhibits variations in performance when detecting clamps of different sizes, indicating sensitivity to size differences. Objects with challenging top surfaces and limited potential graspable areas, such as the mustard, achieved the lowest scores. 
\vspace{-7pt}

\subsection{Other Insights}
From the extensive experiments, we observe that, in simpler environments, the analytical algorithms that consider grasp geometry or grasp physics tend to outperform data-driven models. However, in more complex situations with uneven surfaces or multiple objects, the learning-based approaches may perform better. Opting for the ZED 2i camera increased the performance of the algorithms due to the lower sensor noise. Moreover, we have identified other interesting factors that influence the performance. These factors encompass the weight, friction, and shape properties of the objects being grasped. Even when the algorithms generate satisfactory grasp outputs, objects may still slip due to their physical properties, i.e., weight and surface friction. This trend was evident for objects featuring curved or challenging top surfaces with low friction, like strawberries and pears, as well as for heavier objects such as mustard. Naturally, the performance of data-driven algorithms can be improved by further training them with data from specific operating conditions (e.g., background texture and lighting) and object types. Additionally, it is important to acknowledge the limitations inherent in simulators, specifically when simulating the grasping of small objects, the Gazebo grasp plugin utilized in our benchmarking tool has exhibited instances of occasional failure in grasping.

\section{Open Source Software and Summary of Experimental Configurations}
Our benchmarking software\footnote{ \href{https://github.com/vinayakkapoor/vision_based_grasping_benchmarking}{https://github.com/vinayakkapoor/vision\_based\_grasping\_benchmarking}} is publicly available to facilitate the replication of our experiments. We provide a single setup script that installs all the dependencies in a separate virtual environment, appropriately sets up and builds the workspaces, pulls the weights for the learning-based grasping algorithms, and builds everything in place. While all the experiments of this paper are executed with ROS 1 Neotic, we also included a ROS 2 version of the benchmarking framework. We provide Docker containers for both ROS 1 and ROS 2 to deliver platform independence. The parameters of the experiments can easily be changed via parameter files. In addition, we included templates to easily integrate new grasping algorithms for expanding this benchmarking study. We further provided instructions in our README file of our GitHub repository.

While the details of the experimental setup are given across several sections above, we would like to summarize some of its key properties to facilitate its replication. In our experiments, the camera's viewpoint is directly observing the scene from a top-down viewpoint and at a $0.8$ meters distance. Assuming the camera's image center as the center of the workspace, the objects are placed in six different poses following the instructions in \cite{bib3} with parameters r = 25cm and $\theta$ = 90deg. We used a plain background and a printed image of a wooden background (the image is provided in the GitHub repository). We conducted experiments with two different lighting conditions, 340 lux and 60 lux, ensuring their consistency by measuring them with a digital illumination meter.

\section{Conclusion and Future Expansion}
In conclusion, this study provides valuable insights into the performance of vision-based grasping algorithms. We analyzed the performance differences between analytical approaches and data-driven techniques under various experimental conditions. This research serves as a baseline for benchmarking vision-based grasp synthesis algorithms, and the study can be easily expanded to other algorithms to achieve systematic side-by-side comparisons. To foster further research and standardization, we have made our experiment videos publicly available and open-sourced the benchmarking software used for the experiments.

While this work focused on single-object grasping scenarios that utilize top-down camera viewpoints, we also provided functionality to enable experimentation for different scenarios using the same pipeline. Specifically, we added the option of changing the camera viewpoint on demand using a ROS topic so that experiments can be conducted using different viewpoints if desired. We added a cyclic grasping routine such that the procedure is run until all the objects on the table surface are cleared. This allows the utilization of the pipeline for multi-object scenes. Furthermore, we integrated a popular 6-DOF grasping algorithm, Grasp Pose Detection (GPD) \cite{ten2017grasp}, into our pipeline as a demonstration of utilizing the pipeline with 6-DOF algorithms as well. The ways of utilizing these functionalities are provided in our GitHub documentation. These functionalities enable the study to be extended to a wide range of grasping applications.

\bibliographystyle{IEEEtran}
\bibliography{references}

\newpage
\vfill
\end{document}